\pdfoutput=1
\documentclass[letterpaper, 10 pt, conference]{ieeeconf}  

\IEEEoverridecommandlockouts                              

\usepackage{graphicx}
\usepackage{amsmath}
\usepackage{amsfonts}
\usepackage{multirow}
\usepackage{array}
\usepackage{epstopdf}
\newcolumntype{P}[1]{>{\centering\arraybackslash}p{#1}}
\newcolumntype{M}[1]{>{\centering\arraybackslash}m{#1}}
\usepackage{caption}
\usepackage{subcaption}
\usepackage{amssymb}
\usepackage{bm}
\usepackage{color}
\usepackage{cite}

\title{\LARGE \bf
Pedestrian-Robot Interaction Experiments in an Exit Corridor
}

\author{Zhuo Chen, Chao Jiang and Yi Guo
\thanks{This work was partially supported by the US National Science Foundation under Grant CMMI-1527016.}
\thanks{The authors are with the Department of Electrical and Computer Engineering, Stevens Institute of Technology,
        Hoboken, NJ 07030, USA
        {\tt\small \{zchen39,cjiang6,yi.guo\}@stevens.edu}}%
}

\begin{document}

\renewcommand\citeleft{[}
\renewcommand\citeright{]}
\renewcommand\citepunct{, }

\maketitle
\thispagestyle{empty}
\pagestyle{empty}

\begin{abstract}
The study of human-robot interaction (HRI) has received increasing research attention for robot navigation in pedestrian crowds. In this paper, we present empirical study of pedestrian-robot interaction in an uni-directional exit corridor. We deploy a mobile robot moving in a direction perpendicular to that of the pedestrian flow, and install a pedestrian motion tracking system to record the collective motion. We analyze both individual and collective motion of pedestrians, and measure the effect of the robot motion on the overall pedestrian flow. The experimental results show the effect of passive HRI, where the pedestrians' overall speed is slowed down in the presence of the robot, and the faster the robot moves, the lower the average pedestrian velocity becomes.  Experiment results show qualitative consistency of the collective HRI effect with simulation results that was previously reported.  The study can be used to guide future design of robot-assisted pedestrian evacuation algorithms.
\end{abstract}

\section{Introduction}

With the rapid advances in the development of autonomous mobile robot technologies,
the applications of service and socially assistive robots have been woven into people's daily life.
As robots are performing tasks where human beings are involved,
the problem of human-robot interaction (HRI) arises and has received considerable attention during the past decade.
Understanding the effect of HRI on human's behavior, especially in human and robot collective motion,
is of great significance for robot control or decision-making.
In one of the recent works, HRI is studied and considered for human-aware robot navigation,
where traditional motion planning approach is amended to respect the effect of HRI on human behavior in the presence of mobile robot \cite{kruse2013human}.
In addition to considering human being as moving obstacles,
criteria such as human comfort \cite{luber2012socially,macharet2013learning},
natural motion \cite{bennewitz2005learning,henry2010learning} and socially-adaptive motion \cite{pandey2009framework,kim2016socially}
are taken in to account for robot motion planning.

In the studies of robot-assisted guidance where multiple robots work collaboratively to escort a group of people,
HRI is commonly modeled as repulsive or attractive force that is embedded in the social force model
which governs human motion dynamics \cite{garrell2010local}.
The robot performs either as a leader who directs a group of people towards the destination,
or as a shepherd who regroups people escaping from the group formation.
Thus, human's motion is controlled through the effect of attractive and repulsive force from the leader and shepherd robot, respectively.
In \cite{ferrer2013robot}, the parameters of the social force based HRI model were found by analyzing the data obtained from real-world experiment using an interactive learning approach.

HRI is also studied for robot motion planning in large-scale human crowds.
It has been found interesting how implicit interaction between humans and mobile robots affects human collective motion,
so that desired human crowd behaviors can be achieved by adjusting the motion of robot in human crowd.
Specially, inspired by self-organized phenomena in human collective motion,
attempts have been made to improve human flow efficiency by introducing mobile robots in human crowds.
The large-scale dynamics of human crowds can be modified as human crowds implicitly interact with the robot \cite{kirkland2003simulation,eldridge2005using}.
Previous study \cite{yamamoto2013control} has found that placing a mobile robot in crossing human flows
can help to create diagonal strip pattern in human flow as the human agents try to avoid the collision with the robot.
The strip pattern is a desirable phenomenon for preventing congestion in crossing crowds.
By adjusting their motion, the interacting robots can improve human flow efficiency under different flow densities.
In our early work \cite{jiang2016robot}, simulations based on social force models were carried out to characterize HRI in an uni-directional exit corridor.
An interacting robot was placed and programmed to move in a direction that is perpendicular to that of the flow direction
to regulate the flow velocity in an evacuation situation in a uni-directional corridor.


The focus of this paper is to empirically study passive HRI in an exit corridor for the purpose of robot-assisted pedestrian flow regulation. We conduct real robot experiments with a group of pedestrians walking towards an exit door, and  measure the effect of robot motion on the overall pedestrian flow. Our experimental results show that in an exit corridor environment, a robot moving in a direction perpendicular to that of the uni-directional pedestrian flow can slow down the uni-directional flow, and the faster the robot moves, the lower the average pedestrian velocity becomes. Furthermore, the effect of the robot on the pedestrian velocity is more significant when people walk at a faster speed. This empirical evidence is qualitatively consistent with simulation results reported in our early paper \cite{jiang2016robot}, and can be used for future development on robot-assisted pedestrian flow regulation.

The rest of the paper is organized as follows: Section \ref{Sec_sfm} introduces the the passive pedestrian-robot interaction studied in the paper, and present the experimental setup and procedure. Section \ref{exp_rslt} presents the experimental results, and discuss comparisons with simulation results reported before. The conclusion and future work are presented in Section \ref{conclusion}.

\section{Passive Pedestrian-Robot Interaction and Experimental Setup}\label{Sec_sfm}

\subsection{Passive Pedestrian-Robot Interaction}

\begin{figure}[!t]
    \includegraphics[width=0.4\textwidth]{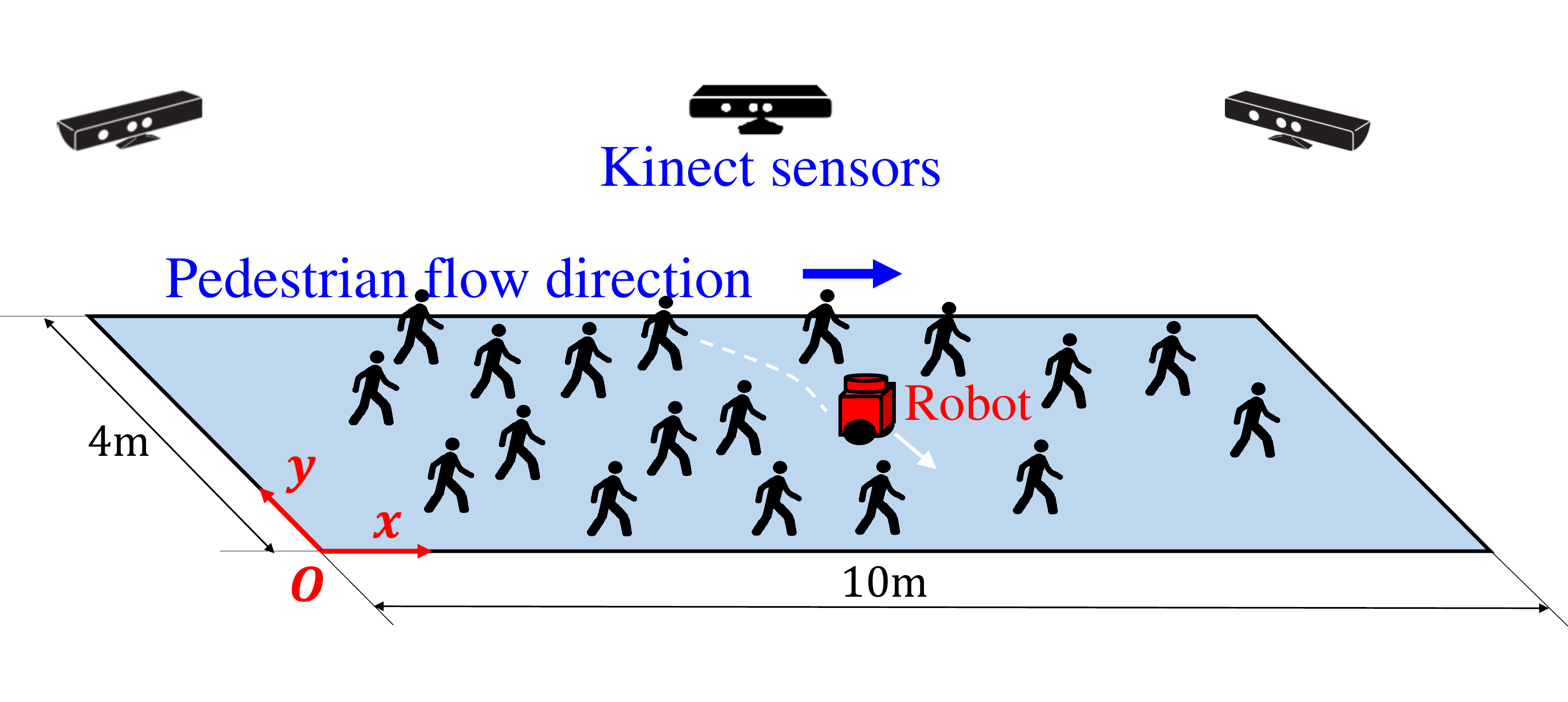}
    \caption{The schematic diagram of pedestrian flow in an indoor corridor with robot-assisted regulation}
    \label{fig:scene}
\end{figure}


We empirically study the regulation of human flow velocity in an exit corridor environment using an interacting robot. As shown in Fig. \ref{fig:scene}, a single interacting robot is added to the pedestrian flow, and moves in a direction perpendicular to that of the pedestrian flow with a predesignated speed. Since the robot's movement is in the way of the pedestrians towards the exit, the pedestrians see the robot and avoid potential collisions by either changing the speed or alter the trajectory of their own. The pedestrian flow is regulated through the effect of {\em passive} HRI. That is, the robot moves in a pre-designated trajectory with a predesignated speed, and the pedestrians  avoid potential collisions when passing through the robot thus {\em passively} interact with the robot. Note that we assume normal situations that the pedestrians can clearly see the robot and their movement behaviors are rational. We do not consider panic situations or nonrational behaviors.

Our hypothesis is that the speed of the robot moving in a direction perpendicular to that of a uni-directional pedestrian flow will affect the average speed of the pedestrians going through the corridor. The goal of the study is to test this hypothesis, measure the effect of the passive HRI by adjusting the robot's speed on the pre-designated trajectory, and compare the effect with the case without the robot.

\subsection{Experimental Setup}

\subsubsection{The Environment}

The experiment was conducted in the Griffith Building of Stevens Institute of Technology on June 22, 2016.
A pedestrian tracking system deployed in the building covers a 10 meter by 4 meter tracking area as shown in Fig.~\ref{fig:scene} with an exit door at the end of the area.
The boundaries of the tracking area are marked with red duct tapes. The tracking system consists of 5 Microsoft Kinect sensors and 6 computers connected to a local area network. A pedestrian detection and tracking software OpenPTrack \cite{munaro2016openptrack} is used.


OpenPTrack is a real-time distributed people detection and tracking software capable of utilizing RGB-D images from the Kinect network. OpenPTrack runs on Robot Operating System (ROS) in Ubuntu 14.04.
During the experiment, each of the 5 Kinect sensors is connected to a computer
and the image stream obtained by the Kinect sensor is fed into a OpenPTrack detection program running on the computer.
The OpenPTrack detection program on each computer processes the images and transmits the detection results to a central computer over the local area network.
The central computer fuses the received data and records the tracking results.

\subsubsection{The Mobile Robot}

\begin{figure}[!t]
    \centering
    \includegraphics[width=0.2\textwidth]{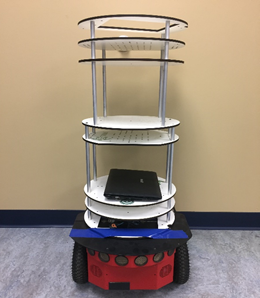}
    \caption{The customized Adept Pioneer P3-DX mobile robot.}
    \label{fig:robot}
\end{figure}

We used a customized Adept Pioneer P3-DX mobile robot, as shown in Fig.~\ref{fig:robot}.
The robot motion dynamics is described by a single integrator model, that is
\begin{equation}\label{robot_motion}
\dot{\bm{x}}_r=\bm{u}_r
\end{equation}
where $\bm{x}_{r}=[x_{r}^x, x_{r}^y]^{T}\subseteq\mathbb{R}^{2}$ denotes the robot position and $\bm{u}_r=[u_{r}^{x},u_{r}^{y}]^T$ denotes robot velocity control. The robot motion is controlled to follow a pre-designed trajectory that is perpendicular to the pedestrian flow direction. The robot velocity in the $y$ direction is controlled to track a sinusoidal signal, and the velocity control in the $x$ direction remains $0$. That is,
\begin{subequations}\label{rob_ctrl}
\begin{equation}
u_{r}^x(t)=0
\end{equation}
\begin{equation}\label{rob_ctrlb}
u_{r}^y(t)=A\omega\sin\left(\omega t \right)
\end{equation}
\end{subequations}
where $A$ is the amplitude of the simple harmonic motion, which is set to 2m (half the corridor width); $\omega$ is the robot motion frequency. This control law makes sure that the robot moves within the boundary of the area, and the upper limit of the controlled speed is $A\omega$.

\subsubsection{Tracking System Calibration}

\begin{figure*}[!htp]
    \centering
    \begin{subfigure}[t]{0.3\textwidth}
        \includegraphics[width=\textwidth]{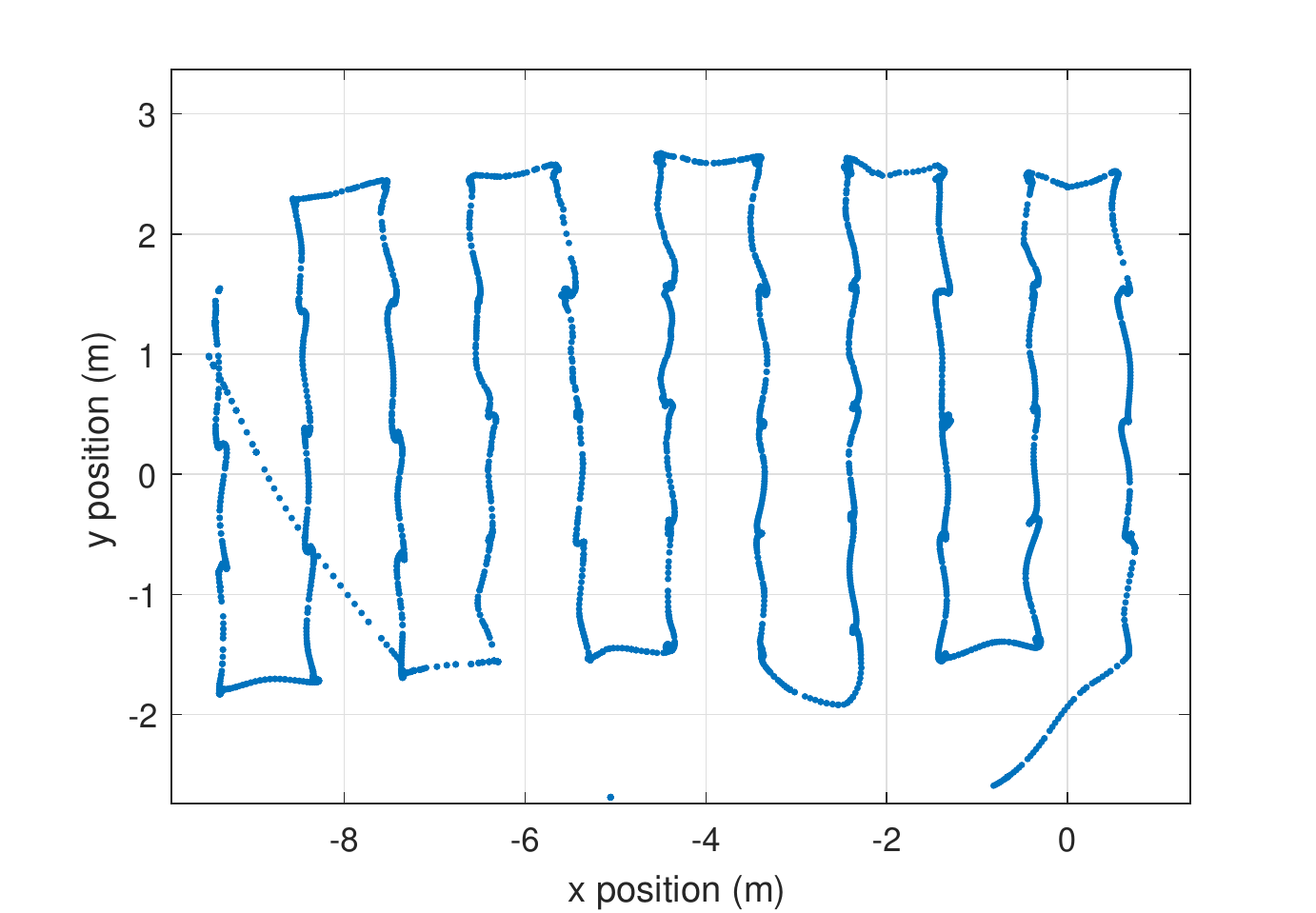}\\
        \caption{}\label{fig_accuracy:walking_pattern}
    \end{subfigure}
    \begin{subfigure}[t]{0.3\textwidth}
        \includegraphics[width=\textwidth]{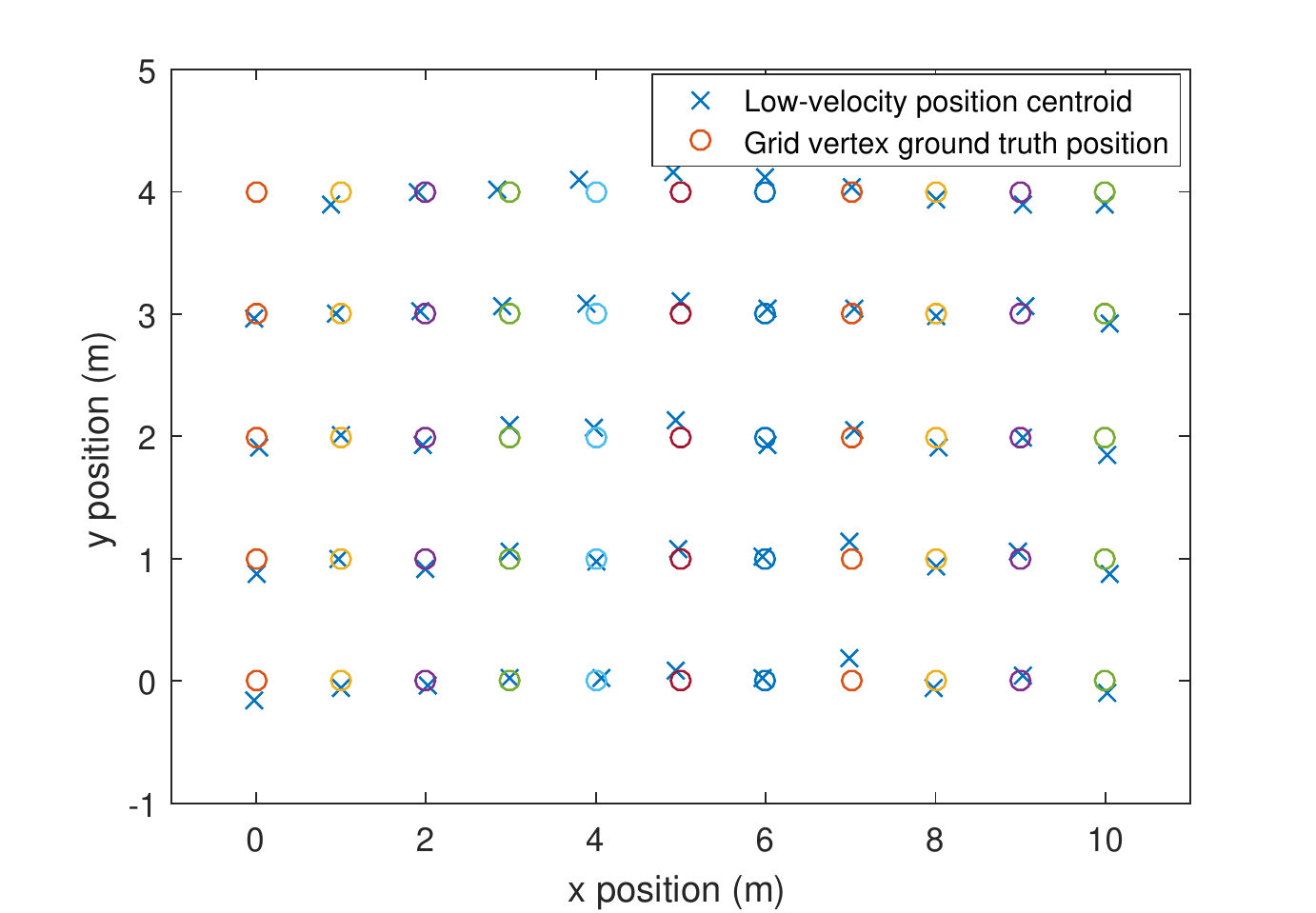}\\
        \caption{}\label{fig_accuracy:compare_measurement_GT}
    \end{subfigure}
    \begin{subfigure}[t]{0.3\textwidth}
        \includegraphics[width=\textwidth]{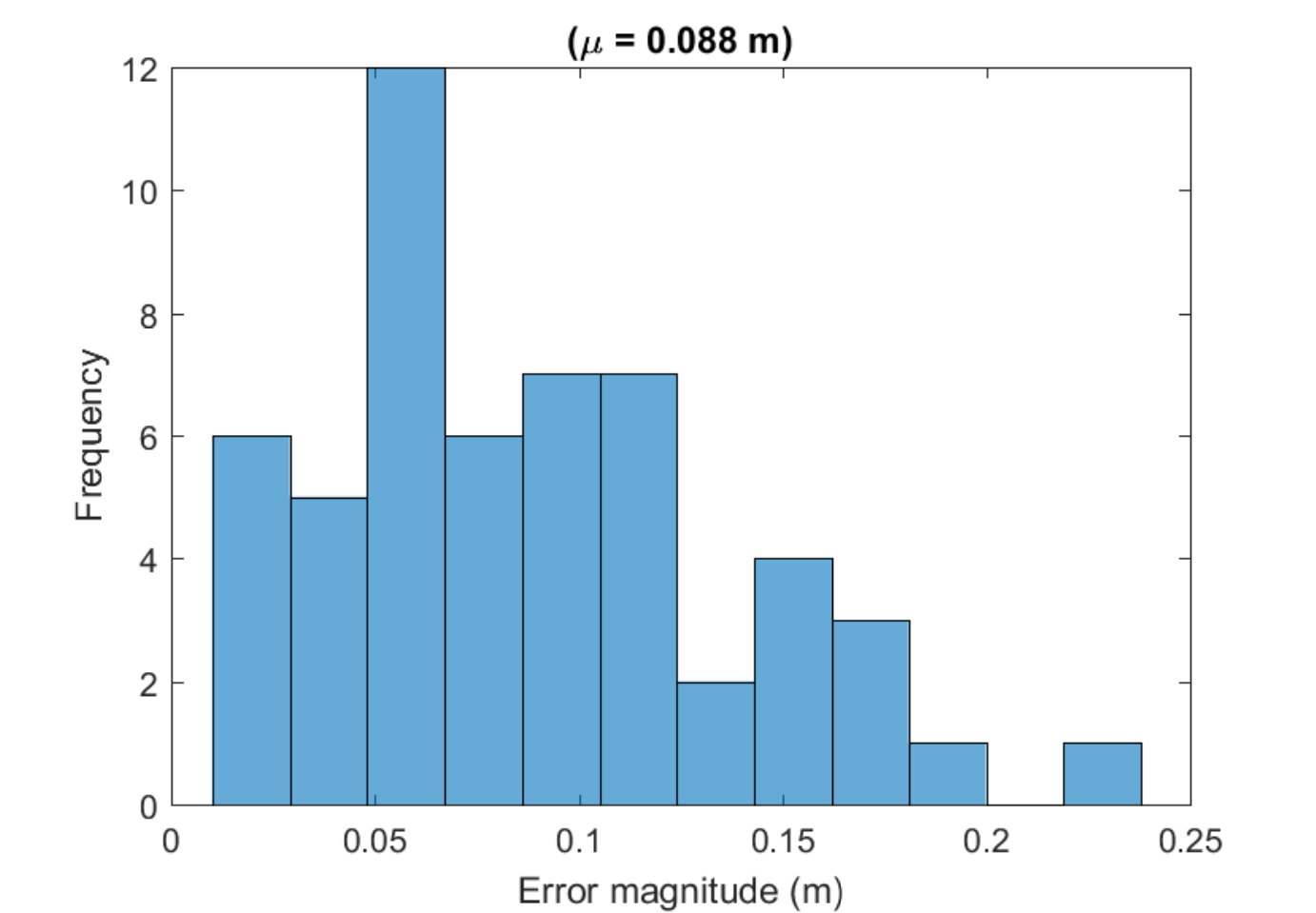}\\
        \caption{}\label{fig_accuracy:hist_clustered_error}
    \end{subfigure}

    \caption{Tracking system calibration: estimation of the coordinate transformation from checkerboard frame to ground frame.
    (a) Walking Pattern in checkerboard reference frame;
    (b) Comparison between measurement and ground truth;
    (c) Measurement error magnitude histogram.
    }\label{fig:accuracy_test}
\end{figure*}

\begin{figure*}[!htp]
    \begin{subfigure}[t]{0.24\textwidth}
        \includegraphics[width=\textwidth]{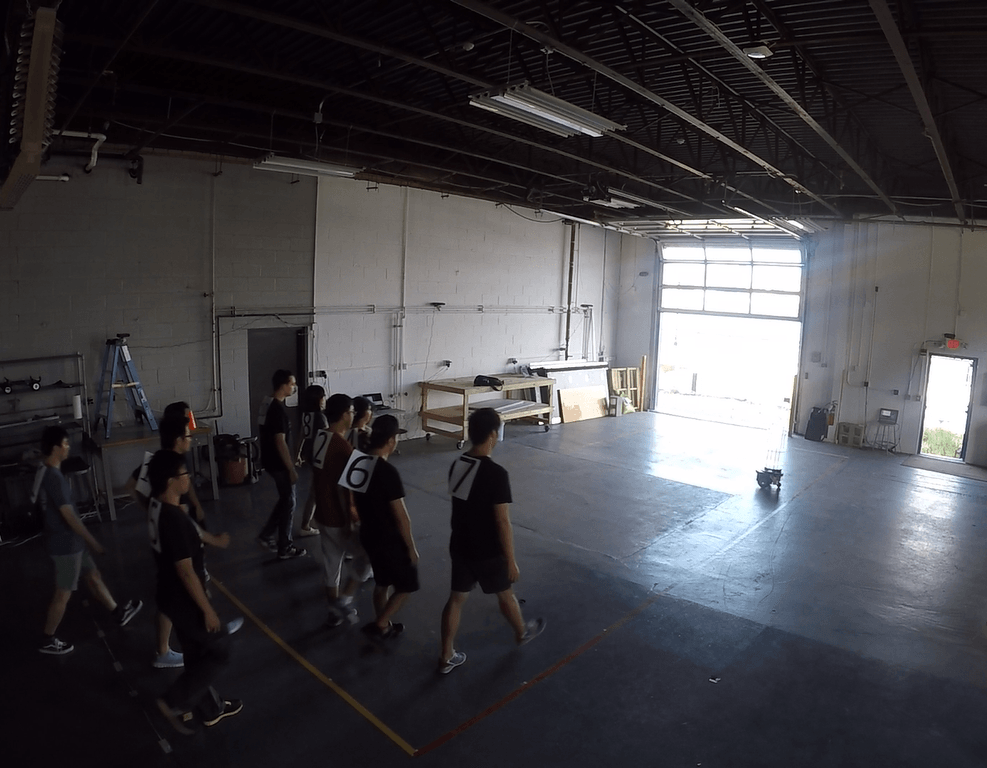}\\
        \caption{t=$4.0$s}\label{fig:video_snapshot1}
    \end{subfigure}
    \begin{subfigure}[t]{0.24\textwidth}
        \includegraphics[width=\textwidth]{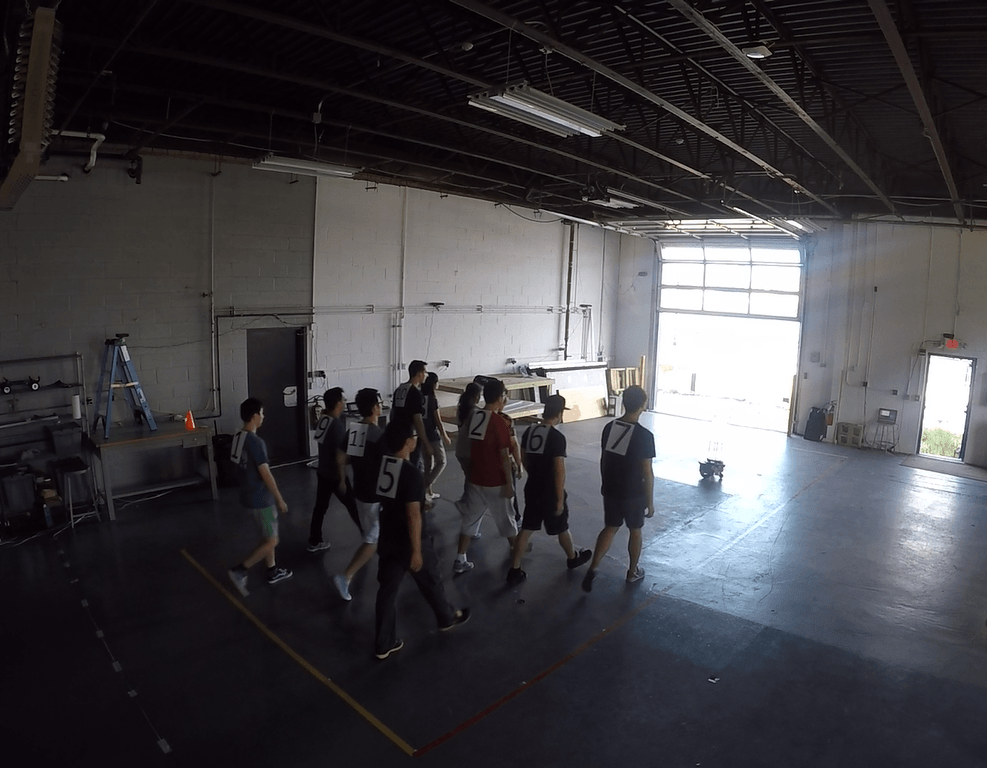}\\
        \caption{t=$6.0$s}\label{fig:video_snapshot2}
    \end{subfigure}
    \begin{subfigure}[t]{0.24\textwidth}
        \includegraphics[width=\textwidth]{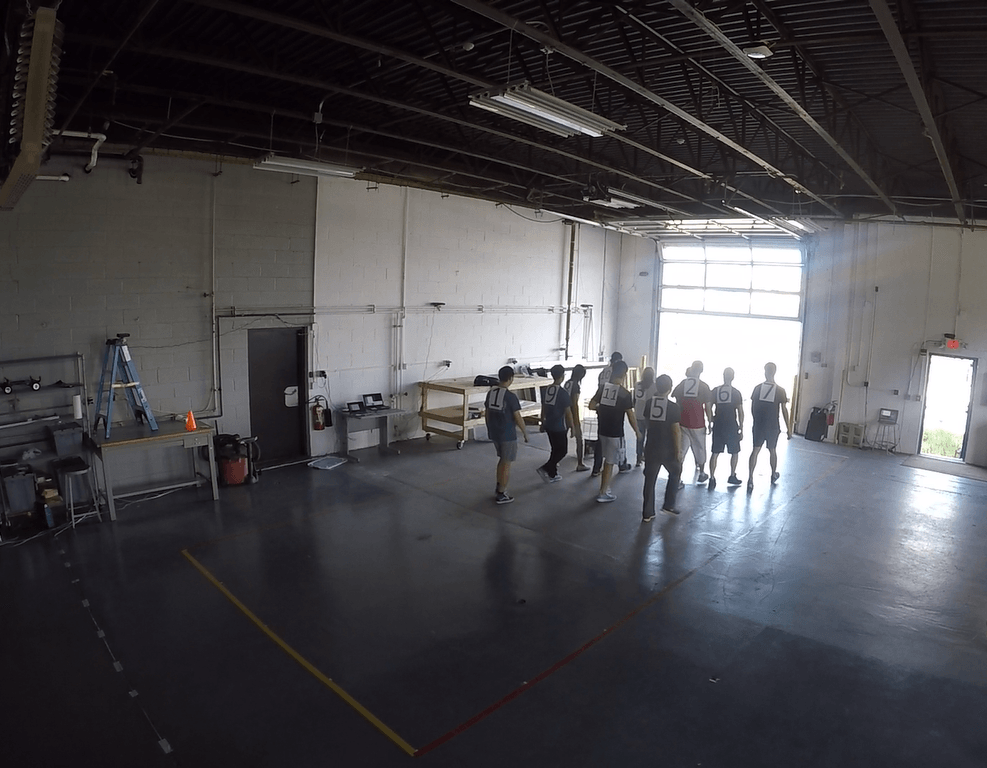}\\
        \caption{t=$9.0$s}\label{fig:video_snapshot3}
    \end{subfigure}
    \begin{subfigure}[t]{0.24\textwidth}
        \includegraphics[width=\textwidth]{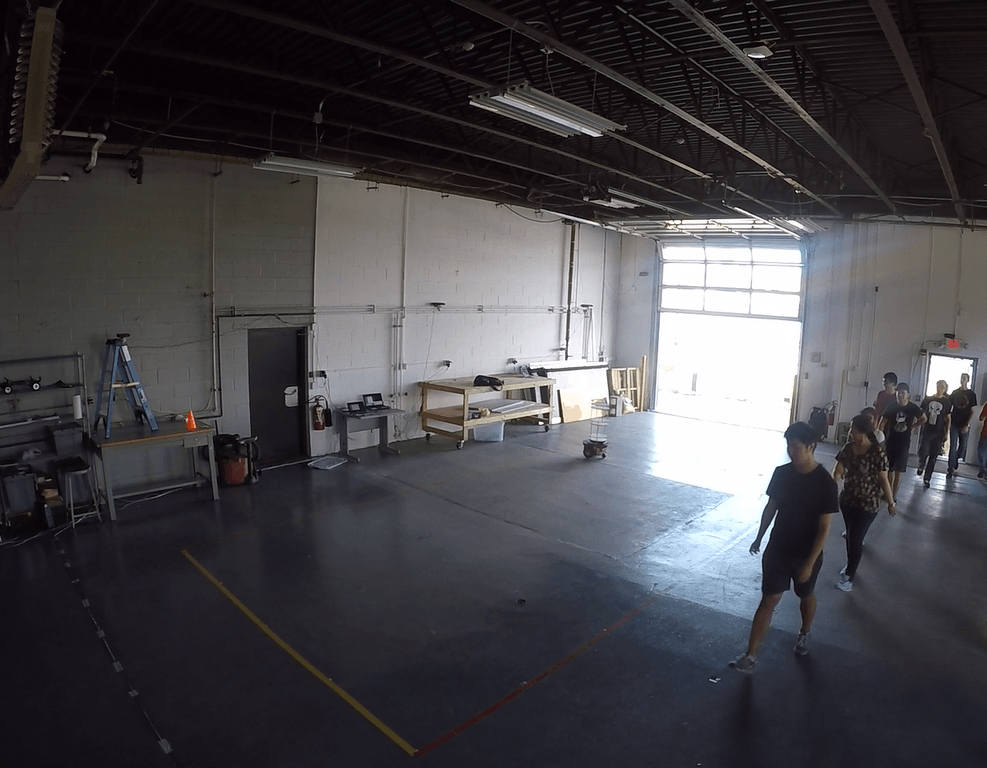}\\
        \caption{t=$24.0$s}\label{fig:video_snapshot4}
    \end{subfigure}
    \caption{Experimental procedure. (a) People start to walk in the bounded area; (b) People walk toward the robot; (c) People interact with the moving robot; (d) People finish one round and walk back to the starting line.}\label{fig:video_snapshot}
\end{figure*}

After the sensors are mounted in place, both intrinsic and extrinsic parameters of the tracking system need to be determined.
Default intrinsic parameters for the Kinect sensors were used without performing any intrinsic calibration.
To calibrate the extrinsic parameters of the tracking system, we printed a 6-by-7 checkerboard with the cell width and height both being 75mm.
The networking capabilities of ROS enable the Kinect sensors' poses to be estimated by:
Step (1) letting a pair of sensors both see the checkerboard to calibrate their relative pose;
Step (2) finding one calibrated sensor and one uncalibrated sensor, and doing Step (1) with this pair of sensors;
Step (3) repeating Step (2) until all sensors are calibrated;
Step (4) placing the checkerboard on the ground where it can be seen by at least 1 sensor to specify the checkerboard reference frame;
and Step (5) performing calibration refinement with people detections \cite{munaro2016openptrack}.

By performing the 5 steps above,
sensor poses are estimated with respect to the checkerboard reference frame defined by the pose of the checkerboard in Step (4),
which does not necessarily coincide with the ground reference frame as shown in Fig.~\ref{fig:scene}.
Here we define the transformation $\mathbf{T}:\mathbb{P}^2(\mathbb{R}) \to \mathbb{P}^2(\mathbb{R})$ that maps the homogeneous coordinates of points in checkerboard reference frame to those in ground reference frame as
\begin{equation}\label{transformation_representation}
    \mathbf{T} =
        \begin{bmatrix}
            \cos \theta & 0 & \Delta x \\
            0 & \sin \theta & \Delta y \\
            0 & 0 & 1 \\
        \end{bmatrix}
\end{equation}
where $(\Delta x, \Delta y)$ represents translation while $\theta$ represents rotation.

To determine the transformation $\mathbf{T}$, the vertices of a unit-distance square grid graph are marked on the tracking area ground.
In Fig.~\ref{fig_accuracy:compare_measurement_GT},
ground truth positions of the 55 vertices marked with circles in Fig.~\ref{fig_accuracy:compare_measurement_GT}
are expressed in checkerboard reference frame as $(x_{\mathrm{gt},k}, y_{\mathrm{gt},k}), \quad k=1,2,...,55$.
A person is asked to walk through every marked vertex of the grid, stop at each mark and remain still for a second or two before proceeding to the next one.
Fig.~\ref{fig_accuracy:walking_pattern} shows an example pattern of trajectory recorded by the tracking system,
starting from $(4,0)$ and ending at $(10,0)$ in ground frame coordinates.

Once the trajectory is recorded,
the measured positions where the subject stops, or low-velocity points, are obtained by filtering out those points in the trajectory whose velocity is greater than $0.2$m/s.
These low-velocity positions are clustered and the centroid of each cluster is marked with a ``$\times$" in Fig.~\ref{fig_accuracy:compare_measurement_GT}.
The low-velocity position centroid expressed in checkerboard reference frame is denoted by $(x_{\mathrm{ct},k}, y_{\mathrm{ct},k}), k=1,2,...,55$.
We solve for the transformation $\mathbf{T}$ by minimizing the sum of squared errors between low-velocity position centroids and their corresponding ground truth positions, that is,
\begin{equation}\label{error}
e\left(\mathbf{T}\right) =
    \sum_{k=1}^{55}
    \left|
    \mathbf{T} \cdot
    \begin{bmatrix}
        x_{\mathrm{ct},k} \\
        y_{\mathrm{ct},k} \\
        0 \\
    \end{bmatrix}
    -
    \begin{bmatrix}
        x_{\mathrm{gt},k} \\
        y_{\mathrm{gt},k} \\
        0 \\
    \end{bmatrix}
    \right|^2
\end{equation}

\subsection{Experiment Procedure}

After the tracking system was set up and calibrated, we conducted experiments with 6 cases as listed in Table~\ref{tab:scenarios}. We compare HRI in two testing scenarios with people walking at a normal speed (Cases \#1, \#2, and \#3) and a faster speed (Cases \#4, \#5, and \#6). In each of the two scenarios, we compare pedestrian motions with: 1) no robot, 2) a slow-moving robot, and 3) a fast-moving robot.  The mobile robot, whose motion is subject to (\ref{robot_motion}), has a pre-set angular frequency $\omega=0.1$rad/s for the slow-moving cases (Cases \#2 and \#5), and $\omega=0.4$rad/s in the fast-moving cases (Cases \#3 and \#6).


In each case, there are 11 participants who have no knowledge of the aim or nature of this experiment. As shown in Table~\ref{tab:scenarios}, in Cases \#1 and \#4, pedestrians walk without robot intervention, while in the rest of the cases, a mobile robot whose motion is subject to (\ref{robot_motion}) is introduced in the environment. The participants are instructed to walk within the boundaries of the tracking area at two different walking speed, i.e., normal or faster. As shown in Fig. \ref{fig:video_snapshot}, the participants waiting on the left side of the tracking area receives an instruction to walk, and they enter the tracking area. Once they exit the tracking area from the right side of the corridor, they walk back through another pathway next to the tracking area and get ready for the next run. Each case consists of 3 repeated runs. Fig.~\ref{fig:video_snapshot} shows the snapshots of a complete run of Case \#3.

{\small
\captionof{table}{Experimental scenarios and parameters}\label{tab:scenarios}
\noindent\makebox[0.48\textwidth]{
\begin{tabular}{|M{1cm}|M{1cm}|M{1.2cm}|M{1cm}|M{1.2cm}|}
        \hline
        \multirow{2}{*}{Case \#} & \multicolumn{2}{c|}{Human} & \multicolumn{2}{c|}{Robot} \\ \cline{2-5}
        & \# of People & Instructed Walking Speed & Angular Freq. $\omega$(rad/s) & Start Position \\ \hline
        1 & 11 & Normal & N/A & N/A \\ \hline
        2 & 11 & Normal & 0.1 & (6.6, 0) \\ \hline
        3 & 11 & Normal & 0.4 & (6.6, 0) \\ \hline
        4 & 11 & Faster & N/A & N/A \\ \hline
        5 & 11 & Faster & 0.1 & (6.6, 0) \\ \hline
        6 & 11 & Faster & 0.4 & (6.6, 0) \\ 
      \hline
\end{tabular}
}}\\

We present the results from collected data in the next section.

\section{Experimental Results}\label{exp_rslt}

In this section, we present the observed HRI behaviors, summary statistics for collected HRI, and compare with numerical simulation results reported in our previous work \cite{jiang2016robot}.

\subsection{Individual HRI Behaviors}

\begin{figure*}[!htp]
    \begin{subfigure}[b]{0.24\textwidth}
        \includegraphics[width=\textwidth]{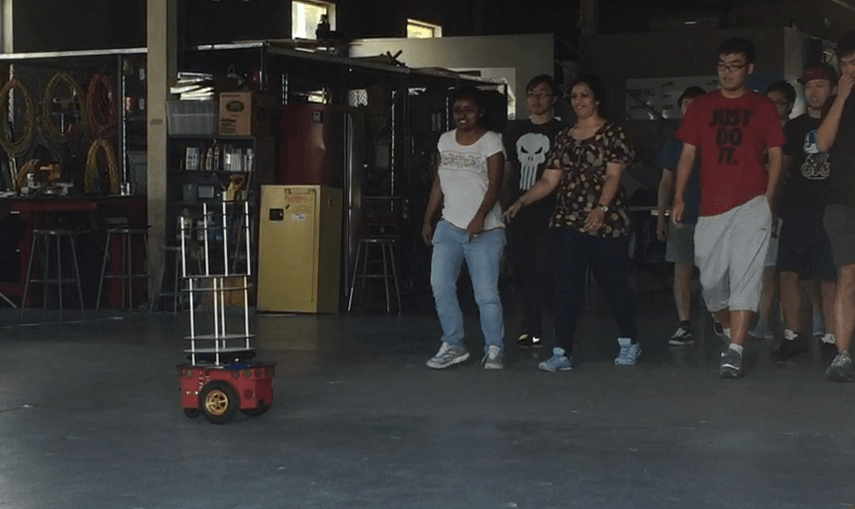}\\
        \caption{$t$=46.0s}\label{fig:video_snapshot_case2_run2_1}
    \end{subfigure}
    \begin{subfigure}[b]{0.24\textwidth}
        \includegraphics[width=\textwidth]{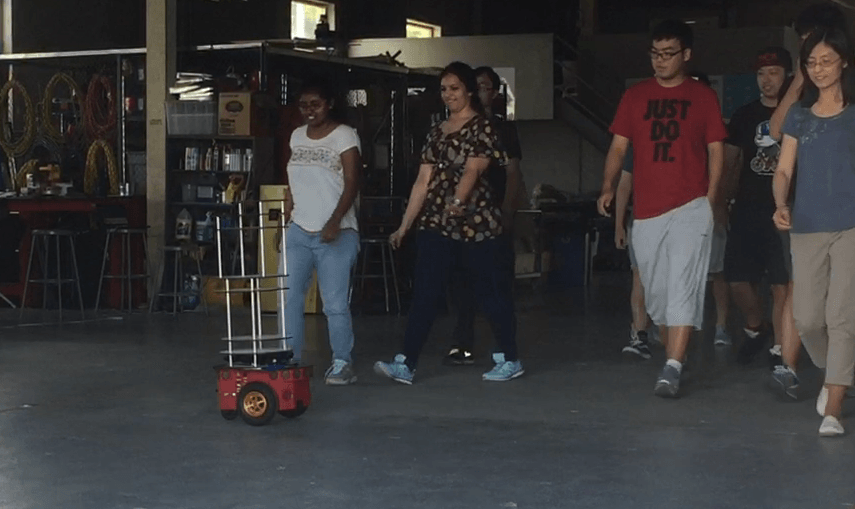}\\
        \caption{$t$=47.0s}\label{fig:video_snapshot_case2_run2_2}
    \end{subfigure}
    \begin{subfigure}[b]{0.24\textwidth}
        \includegraphics[width=\textwidth]{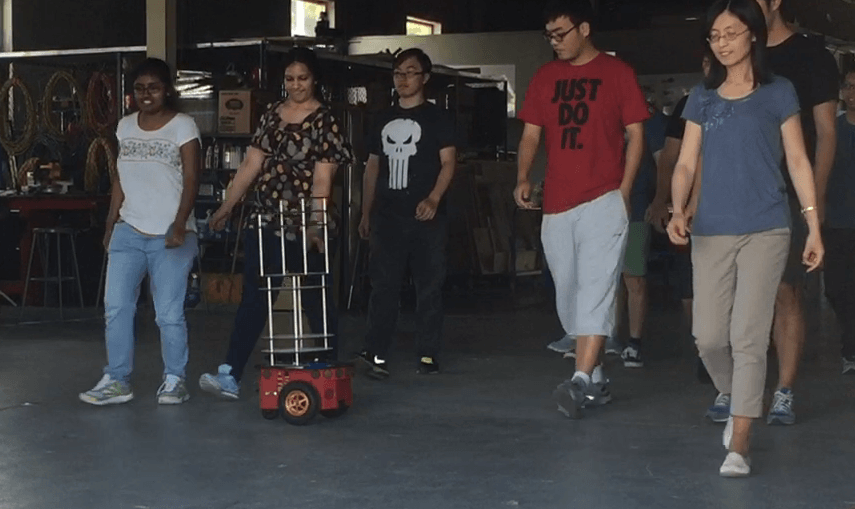}\\
        \caption{$t$=48.0s}\label{fig:video_snapshot_case2_run2_3}
    \end{subfigure}
    \begin{subfigure}[b]{0.24\textwidth}
        \includegraphics[width=\textwidth]{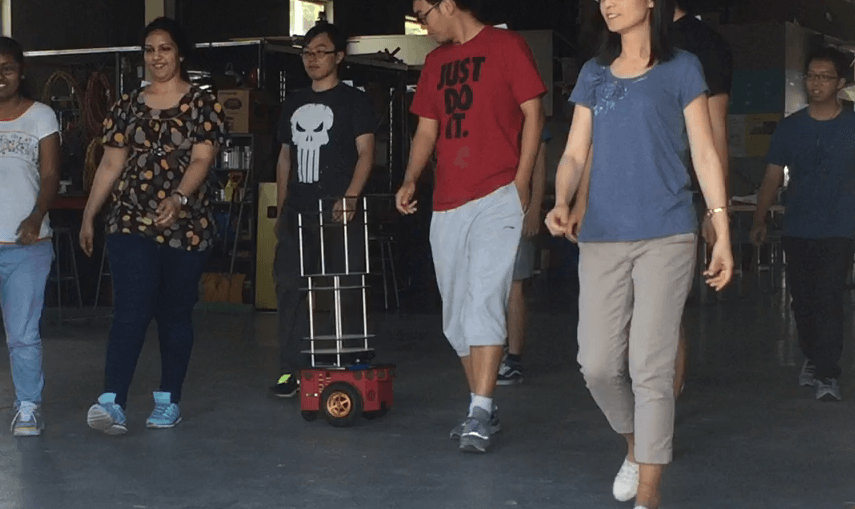}\\
        \caption{$t$=49.0s}\label{fig:video_snapshot_case2_run2_4}
    \end{subfigure}
    \caption{HRI behavior with normal walking speed of pedestrians (Case \#2).
    }\label{fig:video_snapshot_case2_run2}
\end{figure*}

\begin{figure*}[!htp]
    \begin{subfigure}[b]{0.24\textwidth}
        \includegraphics[width=\textwidth]{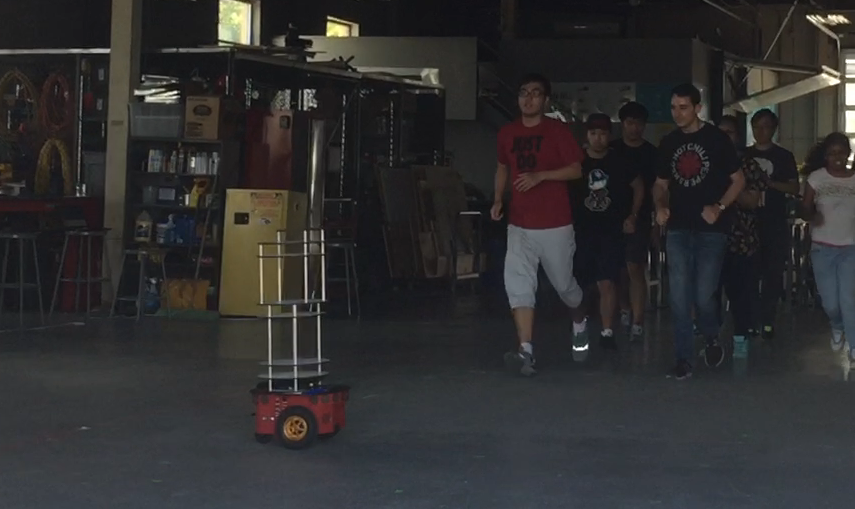}\\
        \caption{$t$=39.0s}\label{fig:video_snapshot_case5_run2_1}
    \end{subfigure}
    \begin{subfigure}[b]{0.24\textwidth}
        \includegraphics[width=\textwidth]{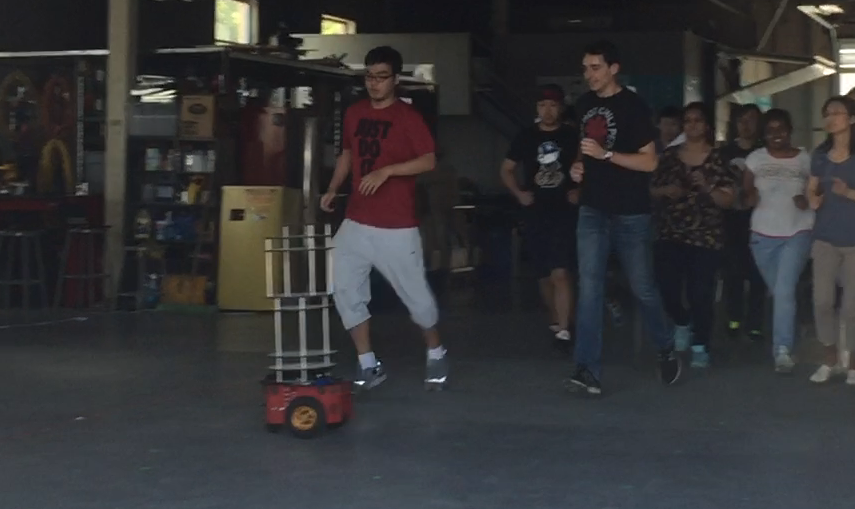}\\
        \caption{$t$=39.6s}\label{fig:video_snapshot_case5_run2_2}
    \end{subfigure}
    \begin{subfigure}[b]{0.24\textwidth}
        \includegraphics[width=\textwidth]{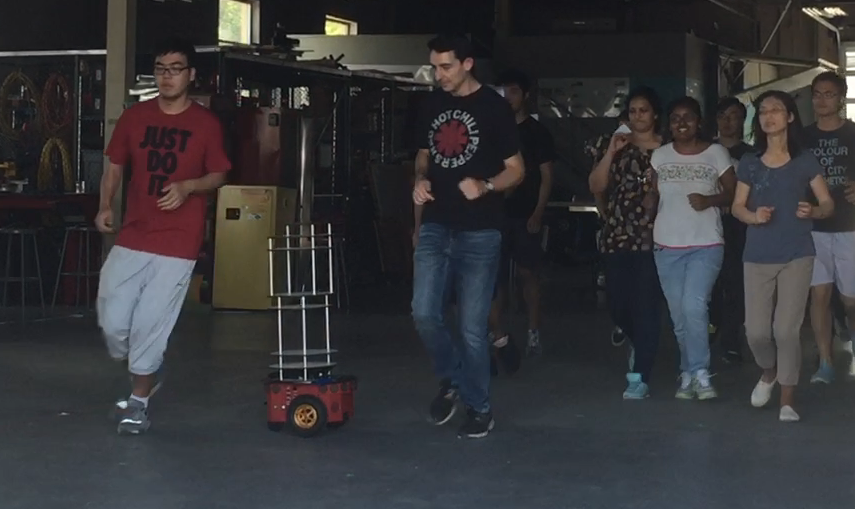}\\
        \caption{$t$=40.2s}\label{fig:video_snapshot_case5_run2_3}
    \end{subfigure}
    \begin{subfigure}[b]{0.24\textwidth}
        \includegraphics[width=\textwidth]{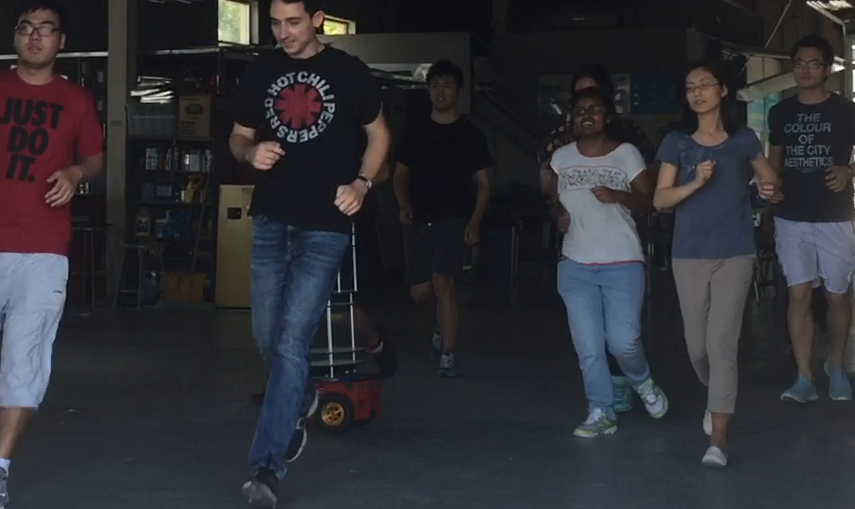}\\
        \caption{$t$=40.8s}\label{fig:video_snapshot_case5_run2_4}
    \end{subfigure}
    \caption{HRI behavior with fast walking speed of pedestrians (Case \#5).
    }\label{fig:video_snapshot_case5_run2}
\end{figure*}

\begin{figure*}[!htp]
    \centering
    \begin{subfigure}[b]{0.28\textwidth}
        \includegraphics[width=\textwidth]{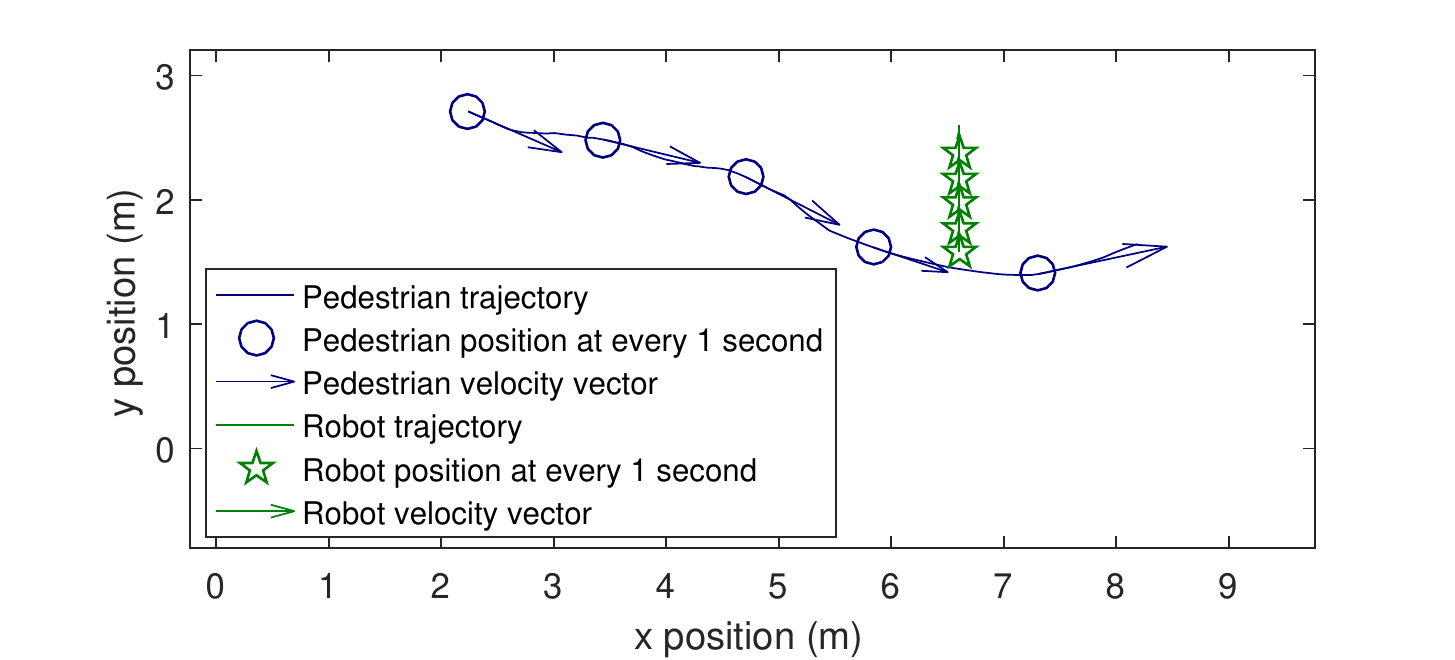}
        \caption{}\label{fig:traj1988}
    \end{subfigure}
    \begin{subfigure}[b]{0.28\textwidth}
        \includegraphics[width=\textwidth]{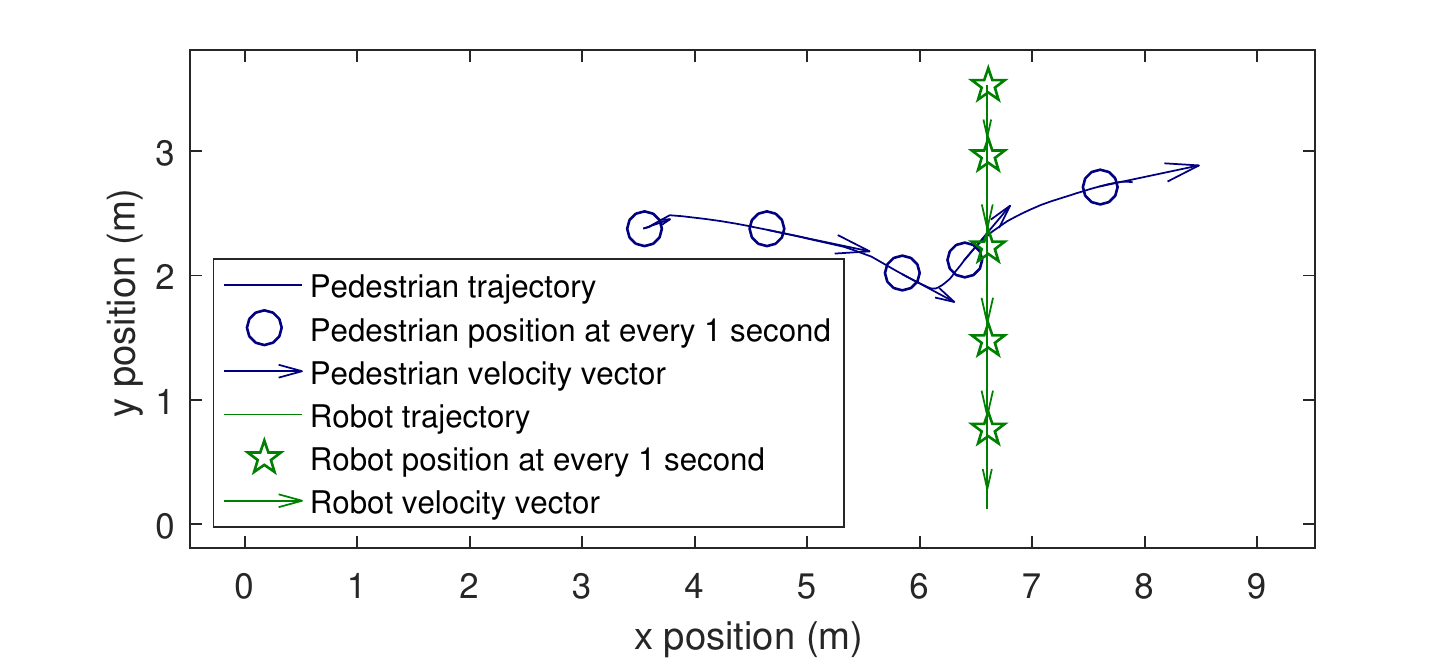}
        \caption{}\label{fig:traj2266}
    \end{subfigure}
    \begin{subfigure}[b]{0.28\textwidth}
        \includegraphics[width=\textwidth]{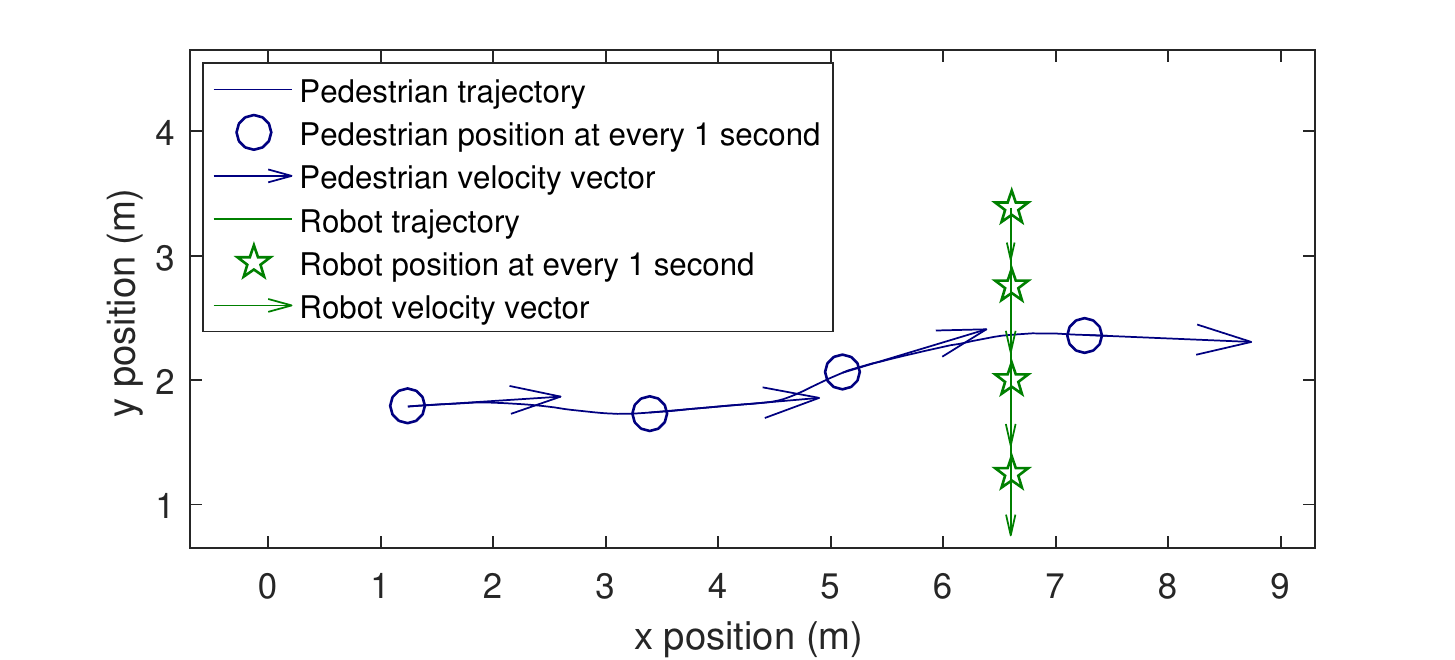}
        \caption{}\label{fig:traj2670}
    \end{subfigure}
    \caption{Single pedestrian and robot interacting trajectories. (a) Case \#2, run \#2, pedestrian \#10, $t$ is from $47.1$s to $51.1$s; (b) Case \#3, run \#2, pedestrian \#1, $t$ is from $40.0$s to $44.0$s; (c) Case \#6, run \#2, pedestrian \#7, $t$ is from $36.7$s to $39.7$s.}\label{fig:single_traj1}
\end{figure*}

We have recorded pedestrian trajectories using the OpenPTrack software and a video recording system. We observe the HRI behaviors when people get close to the robot. It appears that the pedestrian motion behavior changes according to their anticipation of potential collision with the robot, and the density of the area and whether there's room to move around. We observe that the speed change of the robot affects humans' motion. We first describe individual HRI behaviors in this subsection, and the HRI effect on collective motion and the average flow velocity is presented in the next subsection.

Fig.~\ref{fig:video_snapshot_case2_run2} shows the temporal sequence of experiment snapshots in Case \#2, where the pedestrians walk at normal speed.
It can be seen that the pedestrians approach the interacting robot at $t=46$s and start to avoid the robot at $t=47$s.
At $t=48$s and $t=49$s, some pedestrians adjust their walking directions to avoid collision with the robot. Fig.~\ref{fig:video_snapshot_case5_run2} shows the temporal sequence of experiment snapshots in Case \#5, where the pedestrians walk at faster speed.

To understand the motion behavior of pedestrians while interacting with the robot,
we further investigate the change of trajectory and velocity of individual pedestrians.
Fig.~\ref{fig:traj1988} and Fig.~\ref{fig:traj2266} show the trajectories of the pedestrians and the interacting robot in Case \#2 and Case \#3, respectively.
The circles and stars represent the position of the pedestrian and the robot at every 1 second.
The arrow represents the velocity with the length proportional to the velocity magnitude.
From Fig.~\ref{fig:traj1988}, one can see that the pedestrian adjusts his/her walking direction at around $x=4.5$m to avoid the robot as he/she foresees potential collision with the robot.
Meanwhile, the velocity of the pedestrian changes slightly.
One can see from Fig.~\ref{fig:traj2266} that the pedestrian adjusts his/her walking direction at around $x=4.6$m and $x=5.8$m respectively, and meanwhile slows down to avoid collision with the robot.
Fig.~\ref{fig:traj2670} shows the trajectories of the pedestrian and the interacting robot in Case \#6, from which one can see the pedestrian adjusts his/her walking direction at around $x=4.8$m.
The results demonstrate that both the trajectory and velocity of the pedestrians can be affected by HRI.

\subsection{Summary Statistics for Collective HRI}

\begin{figure*}[!htp]
    \centering
     \begin{subfigure}[b]{0.27\textwidth}
        \includegraphics[width=\textwidth]{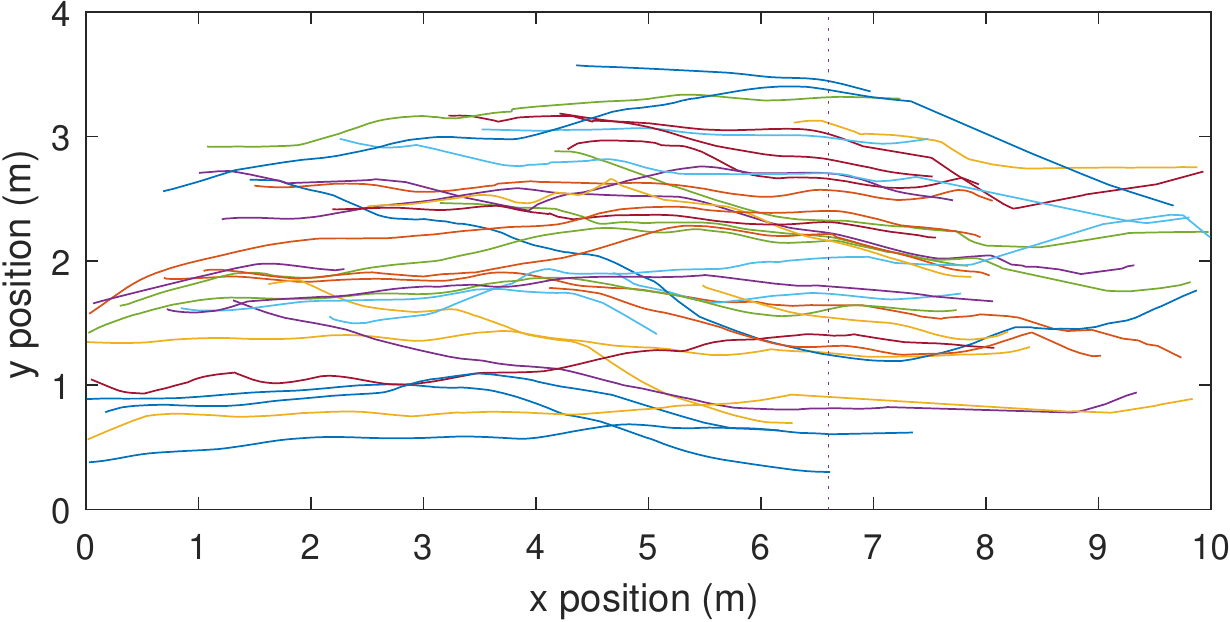}\\
        \caption{}\label{fig_all_traj:trajectories_2}
    \end{subfigure}
    \begin{subfigure}[b]{0.27\textwidth}
        \includegraphics[width=\textwidth]{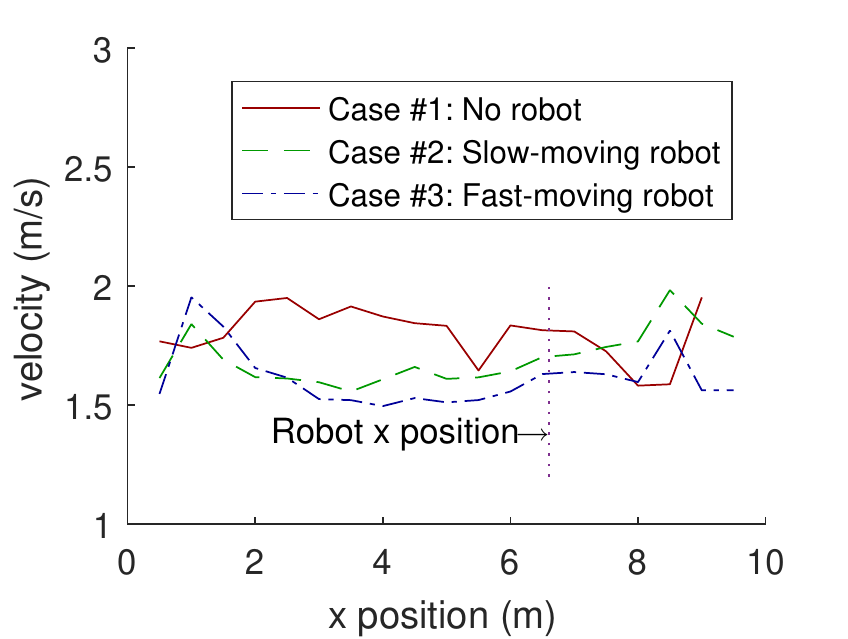}\\
        \caption{}\label{fig:ave_vel_normal}
    \end{subfigure}
    \begin{subfigure}[b]{0.27\textwidth}
        \includegraphics[width=\textwidth]{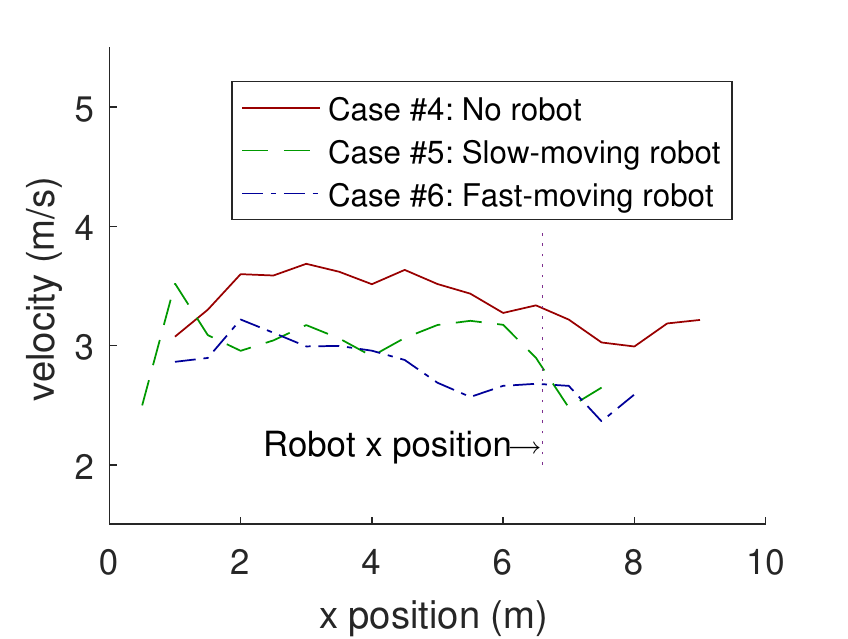}
        \caption{}\label{fig:ave_vel_faster}
    \end{subfigure}
    \caption{(a)Trajectories recorded from all 3 runs of Case \#2. The trajectory followed by each pedestrian is represented by a solid curve. The vertical dotted line denotes the robot trajectory; Average pedestrian velocities vs. x position for (a) normal walking speed of pedestrians, and (b) faster walking speed of pedestrians.}\label{fig:ave_vel}
\end{figure*}

We have collected pedestrian trajectory data for all 6 experimental cases listed in Table~\ref{tab:scenarios}. One set of such trajectories is plotted in Fig.~\ref{fig_all_traj:trajectories_2} for 3 runs of Case \#2. Each data point on the trajectory has a timestamp. The velocity of each pedestrian is calculated by  performing a backward difference on the position signal and then applying a moving average filter to the resulting velocity signal.

To understand the effect of HRI on the collective motion of pedestrians, we plot the average pedestrian velocity profile, which is obtained by averaging all individual pedestrians' velocities along the $x$ axis for each case. Fig.~\ref{fig:ave_vel_normal} shows the testing scenario of normal walking speed of pedestrians (i.e., Cases \#1, \#2 and \#3), and Fig.~\ref{fig:ave_vel_faster} shows the case of the fast walking speed (i.e., Cases \#4, \#5 and \#6). We can see that the average pedestrian velocities are lowered with the presence of robot; and the faster the robot moves, the lower the average pedestrian velocity becomes. This phenomenon is more visible around the robot area (i.e., the vertical dashed line of robot positions shown in the figure) than other areas (e.g., the beginning and the end of the pedestrian trajectories), which indicates an HRI region exists around the robot positions.

\begin{figure*}[!htp]
    \centering
    \begin{subfigure}[b]{0.28\textwidth}
        \includegraphics[width=\textwidth]{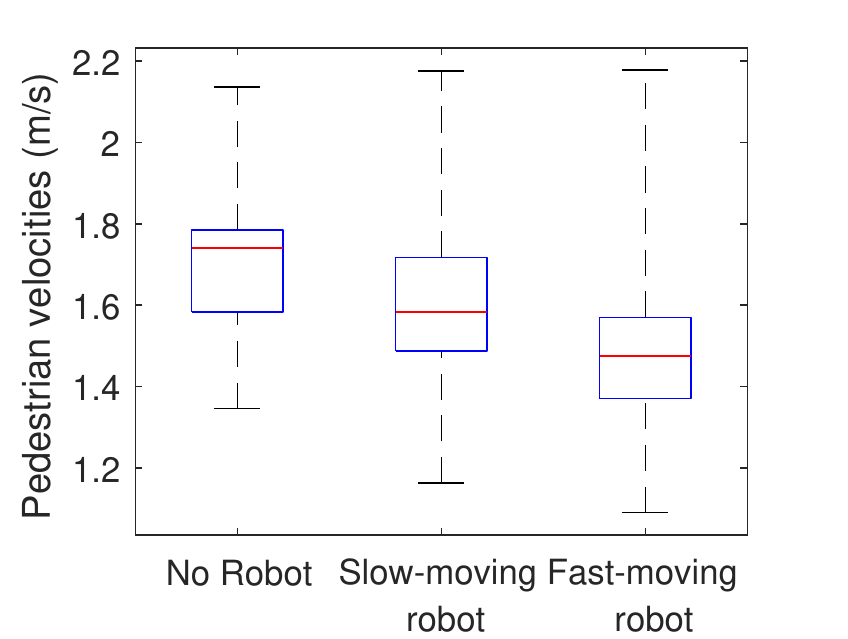}\\
        \caption{}\label{fig:boxplot_normal}
    \end{subfigure}
    \begin{subfigure}[b]{0.28\textwidth}
        \includegraphics[width=\textwidth]{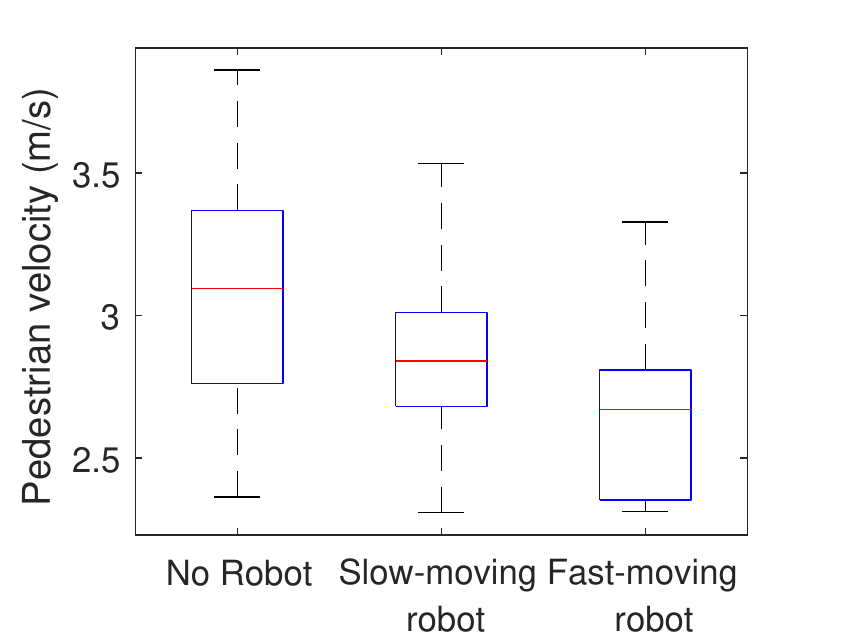}\\
        \caption{}\label{fig:boxplot_faster}
    \end{subfigure}
    \begin{subfigure}[b]{0.28\textwidth}
        \includegraphics[width=\textwidth]{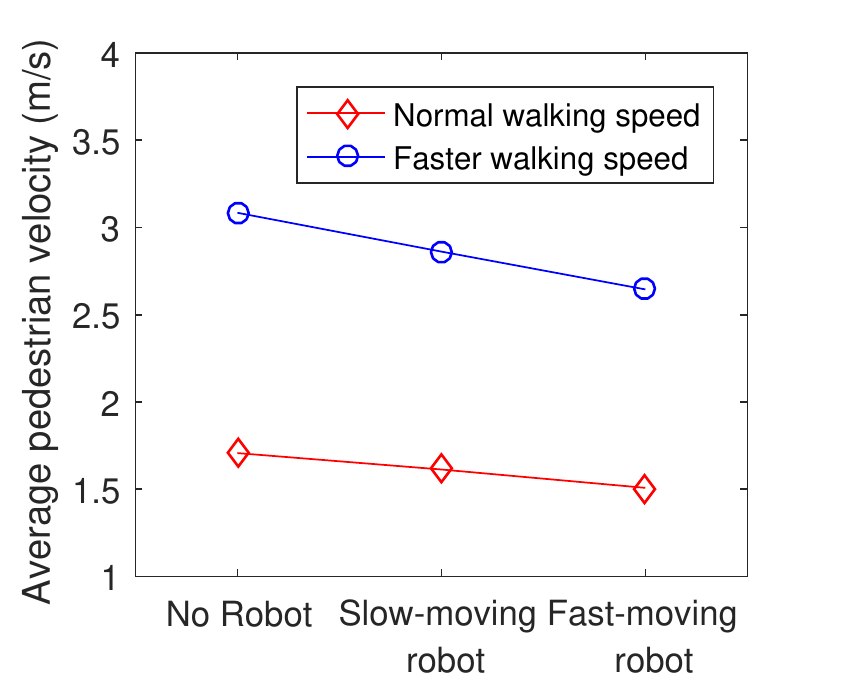}\\
        \caption{}\label{fig:5_va}
    \end{subfigure}
    \caption{Box plot of pedestrian velocities for each case. (a) Normal walking speed; (b) Faster walking speed; (c) Average pedestrian velocity comparison with and without the presence of a moving robot.}\label{fig:3_boxplot}
\end{figure*}

To further understand the collective pedestrian motion, we plot the velocity distribution of pedestrians in each case in the box plot of Fig.~\ref{fig:3_boxplot}, where the central rectangle spans the first quartile to the third quartile, the segment inside the rectangle shows the median, and ``whiskers'' above and below the box show the locations of the minimum and maximum.  We can clearly see the trends of HRI, that is, the first quartile, the median, and the third quartile consistently show that: 1) pedestrian velocities are lowered with the presence of robot, and 2) the faster the robot moves, the lower the average pedestrian velocity becomes.


\subsection{Comparison with Simulations and Discussions}


To compare the experimental results with the numerical simulation results presented in our previous work \cite{jiang2016robot} (Figures 4 and 5 therein), we calculate the pedestrian flow velocity by averaging the velocities over all pedestrians in each case in an HRI region, which is defined the same as in the simulations (i.e., the rectangular regions within 2 meters sideways from the robot path). Fig. \ref{fig:5_va} shows the average pedestrian speed with no-robot and in the presence of a slow- and fast- moving robot in the two cases of normal pedestrian speed and faster pedestrian speed. We can see that the slow-down effect of the pedestrian-robot interaction is more significant for the group of faster-moving pedestrians,  comparing with slower-moving pedestrians. In comparison with simulations reported in our previous work \cite{jiang2016robot}, we can clearly see that the trends of HRI match. Specifically, the following claims are supported by both simulations and our HRI experiments:
\begin{itemize}
\item The pedestrian speeds are affected by the presence of a moving robot;
\item A moving robot slows down a uni-directional flow. The faster the robot moves, the lower the average pedestrian velocity becomes;
\item The effect of the robot on the pedestrian velocity is more significant when people walk at a faster speed.
\end{itemize}

For the tested scenarios, it clearly shows qualitative agreement of HRI in an exit corridor in comparison with numerical simulations based on social force models. The results indicate that it is promising to use a moving robot to regulate pedestrian flows, specifically, slowing down the traffic to a desired average speed. Note that in evacuation scenarios, being able to slow down the traffic and avoid the faster-is-slower effect is often desirable.

With more time and resources, we would have collected more experimental data. For example, we may vary the pedestrian density of the area, and compare the HRI effect with different density setups. Nevertheless, the obtained data can be used to tune simulation models for a better match with real world situations, which will be in the scope of our future research. Also, we plan to conduct future HRI experiments to validate learning-based robot-assisted pedestrian regulation schemes that was presented in our early work \cite{jiang2016robot}.


\section{Conclusion and Future Work}\label{conclusion}

In this paper, we presented a human-robot interaction experiment in an exit corridor.
We investigated the pedestrian behavior when they interact with a robot moving in a direction perpendicular to that of the uni-directional pedestrian flow.
Using the experimental data, we not only studied the individual HRI behavior, but also analyzed collective HRI and the effect of robot on the average pedestrian flow. We  compared the HRI effect on the pedestrian flow with that obtained from numerical simulations based on the social force models that were reported in our early work \cite{jiang2016robot}. We found that the experimental HRI effect on the collective pedestrian flow is qualitatively consistent with numerical simulations. Future work includes calibrating simulation models using experimentally obtained data, and developing learning-based robot motion planner to regulate the collective speed of the pedestrians to a desired level.

\section*{ACKNOWLEDGMENT}
The authors would like to thank the undergraduate students, Peter Smith and Randall Devitt, and the graduate student, Muhammad Fahad, for their assistance in collecting data during the human robot experiments.


\bibliographystyle{ieeetr}
{
\bibliography{ms}
}

\end{document}